\documentclass[final,journal]{IEEEtran}
\usepackage{xcolor}
\usepackage{amsfonts}
\usepackage{amsmath}
\usepackage{mathtools}
\usepackage{multirow}
\usepackage{footnote}
\usepackage{url}
\DeclareMathOperator*{\argminA}{argmin}

\ifCLASSINFOpdf
\usepackage[pdftex]{graphicx}
\usepackage{subcaption}

\begin{document}

\title{Detection of Collision-Prone Vehicle Behavior at Intersections using Siamese Interaction LSTM}

\author{Debaditya Roy,
        Tetsuhiro Ishizaka, 
        C. Krishna Mohan, and
        Atsushi Fukuda
\thanks{D. Roy, T. Ishizaka, A. Fukuda are with the Department of Transportation Systems Engineering, Nihon University, Chiba, 274-8501, Japan. C. Krishna Mohan is with the Department of Computer Science and Engineering, Indian Institute of Technology Hyderabad, India.}
}

\maketitle

\begin{abstract}
As a large proportion of road accidents occur at intersections, monitoring traffic safety of intersections is important. Existing approaches are designed to investigate accidents in lane-based traffic. However, such approaches are not suitable in a lane-less mixed-traffic environment where vehicles often ply very close to each other. Hence, we propose an approach called Siamese Interaction Long Short-Term Memory network (SILSTM) to detect collision prone vehicle behavior. The SILSTM network learns the interaction trajectory of a vehicle that describes the interactions of a vehicle with its neighbors at an intersection. Among the hundreds of interactions for every vehicle, there maybe only some interactions which may be unsafe and hence, a temporal attention layer is used in the SILSTM network. Furthermore, the comparison of interaction trajectories requires labeling the trajectories as either unsafe or safe, but such a distinction is highly subjective, especially in lane-less traffic. Hence, in this work, we compute the characteristics of interaction trajectories involved in accidents using the collision energy model. The interaction trajectories that match accident characteristics are labeled as unsafe while the rest are considered safe. Finally, there is no existing dataset that allows us to monitor a particular intersection for a long duration. Therefore, we introduce the SkyEye dataset that contains 1 hour of continuous aerial footage from each of the 4 chosen intersections in the city of Ahmedabad in India. A detailed evaluation of SILSTM on the SkyEye dataset shows that unsafe (collision-prone) interaction trajectories can be effectively detected at different intersections. 
\end{abstract}


\IEEEpeerreviewmaketitle
\section{Introduction}
Nearly 40\% of all road accidents are recorded at intersections \cite{intersection_crash_report}. Road accidents at intersections can be attributed to a combination of factors like humans (drivers, riders, vehicle occupants, pedestrians, tri-cyclists, and bicyclists), vehicles (design or structure, weight, equipment like seat-belts or tires), and infrastructure or environment (road design, signage, weather, conditions affecting visibility) \cite{roadsafetyreview,statistical1,statistical2}. These factors lead to \textit{black spots} - places where road traffic accidents have historically been concentrated \cite{blackspot1, blackspot2}. Existing research in accident analysis focuses on the identification of black spots through multiple approaches like screening, clustering, and crash prediction \cite{blackspot3}. However, there is no standardized approach that can be followed for all types of roads \cite{blackspot4}. Hence, accident analysis using black spots is impractical for developing countries like India where there is a large disparity in the size of roads and intersections, no earmarked turning lanes, road markings are often blurry and not followed by the drivers, and the lack of enforcement of speed limits. Instead, there is a need to analyze the risk of collisions by monitoring driving behavior.

Driving behavior is affected by navigation around blind spots caused by occlusion of smaller vehicles by larger vehicles, turning distance of different types of vehicles, driver visibility in various environmental conditions, and design of intersections \cite{drivingsurvey}. The effect of the aforementioned factors on  driving behavior manifests in the form of gap distance between vehicles in the same lane and across lanes, acceleration and deceleration of vehicles \cite{drivingsurvey}. Surveillance video cameras can monitor driving behavior effectively over long periods of time like the UA-DETRAC dataset \cite{ua-detrac}. However, they cannot be used to monitor multiple lanes of an intersection simultaneously due to limited field of view and occlusion of vehicles. Aerial videos allow us to monitor all the lanes of an intersection as shown in the VisDrone dataset \cite{datafromsky}. Hence, we design an approach to detect collision proneness at intersections using aerial videos. 

We propose that detection of collision proneness requires the relative distance (distance between the center of two vehicles) and the speed of neighboring vehicles (instantaneous displacement of the center a vehicle between two successive frames) rather than the exact dimensions of the target vehicle, turning radius, the exact distance between vehicles, as used in existing methods for vehicle behavior modeling \cite{social_mixed, social2d,convolutional_social}. This method is particularly useful in case of aerial videos where exact vehicle dimension and turning distance are difficult to obtain for any arbitrary vehicle and intersection. Further, vehicle maneuvers like overtaking, avoiding oncoming traffic, and merging into other lanes occur frequently at intersections. These maneuvers are heavily influenced by neighboring vehicles in the immediate surroundings that leads to gradual or abrupt change in the driving behavior over time. Hence, the state of any vehicle at a particular instance can be expressed based on its relationship with neighboring vehicles.

In the proposed approach, we represent the behavior of every vehicle using the relative distance, speed of the vehicle, and speed information of its neighbors at every frame to form a temporal sequence called \textit{vehicle interaction trajectory}. In literature, Long Short-term Memory (LSTM) networks have been used to represent pedestrian/vehicle spatial trajectories ($x$ and $y$ positions) \cite{social_lstm, social_gan}. In our approach, we encode the vehicle interaction trajectory using the proposed Interaction LSTM module that represents the long-term driving style of a vehicle that is different from existing LSTM based approaches like \cite{social_lstm,social_gan,social_attention} that only consider a small fragment of the trajectories. 

The encoded vehicle interaction trajectories are compared using a Siamese network called Siamese Interaction LSTM (SILSTM) to detect unsafe and safe interaction trajectories. Though SILSTM can learn vehicle interaction behavior, it still needs labels to separate unsafe and safe interaction trajectories. Annotation of unsafe interaction trajectories is challenging and highly subjective in lane-less traffic (drivers do not follow lane-discipline) due to the irregular driving behavior such as a) \textit{staggered following} - following vehicle is staggered with the leader vehicle, b) \textit{non-lane passing} - two-wheelers driving between lanes and passing vehicles in lanes, c) \textit{following between two vehicles} - vehicles occupy any lateral position on roadway for better passing opportunities, d) \textit{multiple leaders} - lane-less movement and different vehicle sizes may cause a vehicle to follow multiple leaders, and e) \textit{lateral movement} - different vehicles have different capability of lateral movement \cite{lanelessbehavior}. To arrive at an objective annotation scheme for unsafe and safe interaction trajectories that adhere to the aforementioned lane-less behavior, we use the characteristics of interaction trajectories from accidents computed by the collision energy model \cite{whoareyouwith}. The interaction trajectories which have similar properties to accident interaction trajectories are labeled as unsafe while the rest are labeled as safe.

The evaluation of collision prone (unsafe) vehicle interaction trajectories requires an aerial dataset of lane-less traffic. However, existing datasets like UA-DETRAC \cite{ua-detrac} and VisDrone \cite{visdrone}  only cover lane-following traffic. Hence, we introduce a new dataset called SkyEye \footnote{Dataset available on request.} in this work to monitor highly heterogeneous traffic with mostly two-wheelers that maneuver between the gaps of large stationary vehicles. In such traffic, the detection and tracking of two-wheelers in lane-less traffic more challenging. Without the detection and tracking information for the individual vehicles, it is even more challenging to detect unsafe driving behavior. So, in our SkyEye dataset, we provide 4,021 annotated vehicle tracks from 4 intersections in the city of Ahmedabad in India to facilitate research in lane-less mixed traffic conditions.

The main contributions of the work are as follows:
\begin{itemize}
    \item An view-independent Siamese interaction LSTM (SILSTM) network for detecting collision-prone vehicle interaction trajectories.
    \item A large annotated aerial dataset called SkyEye for studying lane-less mixed traffic at different types of intersections.
    \item An objective annotation scheme for collision-prone interaction trajectories using the collision energy model.
\end{itemize}

The rest of the paper is organized as follows. Section \ref{related} reviews relevant existing literature. In Section \ref{proposed}, the proposed approach is described in detail, and the evaluation results are presented in Section \ref{experiment}. Finally, the conclusion is presented in Section \ref{conclusion}.

\section{Related Work}\label{related}

In this section, we describe the relevant literature on accident detection. We also discuss interaction modeling as it is integral to accident detection and collision analysis.

\subsection{Accident Detection}
Accident detection in surveillance videos has been studied in literature as either an anomaly detection problem \cite{accident_anomaly3,accident_anomaly2, accident_anomaly1} or vehicle tracking based detection of interactions \cite{anticipating1, cadp, accident_anticipating2}. The reason for treating accidents as anomalies arose due to the unavailability of a large number of recorded accident examples when compared to normal activities. The number of accident examples considered in \cite{accident_anomaly3}, \cite{accident_anomaly2} and \cite{accident_anomaly1} is 8, 6, and 150, respectively. Clearly, such a small number of examples is not sufficient to learn the spatio-temporal dynamics of accidents, especially when there are more than 13 different accident scenarios that are possible \cite{accidenttypes}.  Hence, the anomaly detection based approaches represent regular vehicle motion as: 1) interaction fields \cite{accident_anomaly3}, 2) trajectory features extracted from spatio-temporal video volumes using auto-encoders \cite{accident_anomaly2}, or 3) bag-of-features extracted from 3D convolutional networks (C3D) \cite{accident_anomaly1}. Then the deviation from normal vehicle behavior is used to detect accidents. While this technique can detect abnormal behavior, dense lane-less traffic often results in very close encounters between vehicles at low-speeds that can appear as accidents. 

With more examples of accidents and better vehicle tracking, vehicle interactions during accidents can be represented more accurately. A dataset of 678 dashboard camera videos containing accidents was presented in \cite{anticipating1}. Using a dynamic spatial attention recurrent neural network (DSA-RNN), the authors in \cite{anticipating1} were able to anticipate accidents before they occurred. Owing to a large number of videos, the DSA-RNN was trained to recognize the change in spatial behavior of the vehicles before, during, and after accidents. The spatial representation of the different vehicles was obtained using a spatial attention network based on VGGNet \cite{vgg}. The attention network only focused on the regions that extracted features based on positive detections of an object detector that was trained to recognize vehicles. The trend for larger accident datasets continued with the Car Accident Detection and Prediction (CADP) dataset \cite{cadp} that had 1,416 recorded accidents from surveillance traffic cameras. The authors also demonstrated the ability to anticipate accidents using augmented context mining (ACM) for recognizing smaller objects better with existing object detectors. With ACM, different sized object region proposals were produced based on manual annotation, and the one with the best detection score was retained. Combining ACM based object detection with DSA-RNN \cite{anticipating1}, the authors showed that faster and more accurate accident anticipation could be achieved. 

The largest accident dataset till date called the Near-miss Incident DataBase (NIDB) was introduced in \cite{accident_anticipating2} with 4,594 near-miss incidents recorded from dashcam videos. The authors presented a new loss function called Adaptive Loss for Early Anticipation (AdaLEA) for training RNNs (particularly LSTMs and Quasi-RNNs) that could adaptively change the loss value based how early the network detects an accident before the actual incident. This was a departure from the linear loss proposed in \cite{anticipating1} where the same loss value was used regardless of how early or late an accident was detected. The large number of videos in NIDB helped the authors to pre-train the RNN in order to achieve the earliest prediction of accidents as compared to existing methods. It is important to note that the ability to anticipate accidents depends largely on the identification of every vehicle-vehicle interaction. Hence, we summarize the various interaction modeling methods in the literature.

\subsection{Interaction Modeling}
The most popular method for interaction modeling in traffic flow analysis is the car-following model \cite{car_following} that is used to describe homogeneous traffic with lane discipline. More recently, to accommodate motorcycle-heavy traffic, a tri-class flow (considering bus, car, and motorcycle as separate flows) was empirically studied in \cite{empirical_tri_class}. The traffic flow problem was described as two-wheeler accumulation in different lanes alongside buses and cars, which were segmented as vehicle packets. However, these vehicle packets were still segregated by lanes. Such a packet formation fails to account for the unique kinetic characteristics of two-wheelers riding between lanes as suggested by the authors in \cite{empirical_tri_class}. Hence, interaction models based on social force \cite{original_social_force} were developed to describe vehicle behavior in lane-less traffic \cite{social_mixed, social2d}. Social force models categorize vehicle behavior based on three forces: 1) \textit{attraction} between vehicles moving together as a group, 2) \textit{repulsion} that refers to the minimum distance maintained between members in a group, and 3) \textit{coherence} that means vehicles moving together in a group maintain similar velocity. However, social force models need a large number of parameters to calculate the components of each of these forces for every vehicle-vehicle interaction.

The complexity in defining the social forces explicitly was overcome by methods that learn the relationship between different targets based on the relative distance between their trajectories using Long Short-Term Memory (LSTM) \cite{social_lstm}, Recurrent Neural Networks (RNN) \cite{social_attention}, and Generative Adversarial Networks (GAN) \cite{social_gan}. In all these approaches, the relationship between a target and its neighbors is stored in a shared layer that helps in predicting the future trajectory of the target.  The predictions are then compared with the ground truth, and the errors are used to update the weights in the shared layer as well as the representation layers in the LSTM or RNN. Given an unknown trajectory, these networks also use a part of the trajectory and the shared layer information to generate the future trajectories based on both the distance and probability of collision in the future with neighboring vehicles. However, as the shared layer is learned to produce safe trajectories, it cannot be used to learn the dynamics of accident trajectories. Moreover, the temporal history used to learn the shared layer is designed for processing immediate behavior (between 8 to 12 time-steps) which is not suitable for describing long-term vehicle behavior at intersections involving a hundred or more time-steps. 

\section{Proposed Approach}\label{proposed}
In lane-less traffic, drivers adjust vehicle movements by estimating the motion of neighboring vehicles during overtaking, avoiding oncoming traffic, and merging into other lanes. The neighboring vehicles are in-turn influenced by other vehicles in their immediate surroundings that could lead to a change in their driving behavior over time. Hence, there is a need to effectively represent neighborhood information for every vehicle and the process of representation is described subsequently.

\subsection{Interaction LSTM}
Mathematically, the neighborhood information for any particular vehicle $i$ is represented by the relative distance of the vehicle $i$ with the other vehicles, the speed of vehicle $i$, and the speed of the vehicles in the neighborhood. This neighborhood information which defines the vehicle trajectory is encoded into the input vector $\mathbf{a}^t_i = [d^t_{i1}, d^t_{i2}, \cdots, d^t_{ij}, v^t_i, v^t_1, v^t_2, \cdots, v^t_j]$ for every time step $t$. The distance between the neighboring vehicles $i$ and $j$ at time $t$ denoted by $d^t_{ij}$ and the instantaneous speed for vehicle $i$ at time $t$ is denoted by $v^t_i$. The nearest $j$ neighbors are chosen for representing the state of vehicle $i$ at time $t$. The number of neighbors can be varied to obtain the best possible representation, and we provide ablative studies in Section \ref{experiment} to demonstrate the effect of the same. The neighborhood information obtained for a single frame is concatenated to obtain the interaction trajectory for the entire duration of $N$ frames when the vehicle $i$ is present in the video. Hence, the interaction trajectory sequence for a vehicle $i$ is expressed as $\mathbf{a}_i = [\mathbf{a}^1_i, \mathbf{a}^2_i, \cdots, \mathbf{a}^N_i]$.

Recurrent Neural Networks (RNN) \cite{rnn} can be used to represent the interaction trajectory sequence obtained above. At every time step $t \in \{1, \cdots, N\}$, the hidden state vector $\mathbf{h}_t$ can be updated based on the equation $\mathbf{h}_t = \sigma\left(\mathbf{W}\mathbf{a}^t + \mathbf{U}\mathbf{h}_{t-1}\right)$, where $\mathbf{W}$ is the weight matrix from the input to the hidden-state vector and $\mathbf{U}$ is the weight matrix that links the hidden-state vector from the previous time step $\mathbf{h}_{t-1}$, and $\sigma(.)$ denotes the logistic function.

In dense traffic, vehicles have to ply slowly especially while entering and exiting intersections. Hence, the average length of the trajectory for each vehicle is more than a hundred time-steps (each time-step represents 1/3 of a second). Standard RNNs suffer from vanishing gradient problem in which the back-propagated gradients become extremely small over long sequences. Hence, the LSTM model was introduced \cite{lstm} that sequentially updates the hidden-state representation like an RNN at each time step but alleviates the vanishing gradient problem by introducing three gates for information control to the memory state $\mathbf{s}_t$. The output gate $\mathbf{o}_t$ determines how much of the memory state should be transferred to the next node. The input gate decides the contribution of input $\mathbf{a}_t$ at time-step $t$. Finally, a forget gate $\textbf{f}_t$ is used to control how much of the history of the trajectory should be forgotten. With relation to this work, we refer to this LSTM as an interaction LSTM where each memory state in the LSTM $\textbf{s}_t$ stores a part of the state of a vehicle at time $t$ in terms of the distance and speed of neighboring vehicles. The output gate determines the proportion of information that should be passed across the memory states at each time step. The input gate determines the amount of neighborhood information that should be allowed at time $t$ to update a part of the vehicle state stored in $\textbf{s}_t$ and the forget gate decides how much of the previous vehicle state affects the present vehicle state. 

Every LSTM is parametrized by the input weight matrices and the previous state for each of the gates along with the memory cell. In this work, the LSTMs are formulated with logistic function $\sigma(.)$ on the gates and the hyperbolic tangent ($tanh$) as the activation functions. This formulation can be described mathematically as  
\begin{equation}
\begin{split}
\mathbf{i}_t = \sigma(\mathbf{W}_i \mathbf{a}_t + \mathbf{U}_i \mathbf{h}_{t-1}) \\ 
\mathbf{f}_t = \sigma(\mathbf{W}_f \mathbf{a}_t + \mathbf{U}_f \mathbf{h}_{t-1})  \\
\mathbf{o}_t = \sigma(\mathbf{W}_o \mathbf{a}_t + \mathbf{U}_o \mathbf{h}_{t-1}) \\
\Tilde{\mathbf{s}}_t = tanh(\mathbf{W}_s \mathbf{a}_t + \mathbf{U}_s \mathbf{h}_{t-1}) \\
\mathbf{s}_t = i_t \odot \Tilde{s}_t + \mathbf{f}_t \odot \mathbf{s}_{t-1} \\
\mathbf{h}_t = o_t \odot tanh(\mathbf{s}_t),
\end{split}
\end{equation}
where $\odot$ denotes the element-wise product. The matrices $\mathbf{W}_i$ and $\mathbf{U}_i$, $\mathbf{W}_f$ and $\mathbf{U}_f$, and $\mathbf{W}_o$ and $\mathbf{U}_o$ are associated with the input, output, and forget gates, respectively.

Bidirectional LSTMs were introduced in \cite{bilstm} to incorporate both future and past context of a sequence by using a separate LSTM on the reversed sequence. The output of the combined model at each time step is computed as the concatenation of the outputs from the forward and backward networks. Analyzing the safety of a trajectory at any time step is affected both by the recent past and the immediate future. Particularly, any sudden change in the speed of the vehicle before and after any time step determines the propensity of a collision which can be adequately captured by a bidirectional LSTM. Hence, in this work, we use the bidirectional LSTM models to model the interaction trajectories. Next, we describe the comparison of the modelled trajectories using a Siamese network.

\subsection{Siamese Interaction LSTM}
In dense lane-less traffic, the gap between vehicles is quite narrow at the intersection and similarly when they leave the stop sign. The gradual increase or decrease in vehicle speed for a particular vehicle is also dependent on the volume of vehicles entering or exiting an intersection. These incidents are flagged as unsafe if either a collision energy-based formulation  \cite{social_etiquette} or interaction LSTM architecture described above. As these cases arise naturally out of lane-less traffic, it is essential to classify such interactions from collisions. Hence, we propose a Siamese Interaction LSTM (SILSTM) that compares two trajectories represented using interaction LSTMs to differentiate between safe and collision-prone interaction trajectories in dense lane-less traffic. 

Siamese networks \cite{siameseog} are neural networks with two inputs that share the same weights called tied weights. The outputs of these networks are compared using a distance measure like cosine, Manhattan, or Euclidean distance. The training inputs for a Siamese network consists of two input sequences and a label to indicate whether they are similar or dissimilar. The Siamese network is trained in a way so as to minimize the distance between features of the same class and maximize the distance between dissimilar sequences \cite{siameselstm}.

\begin{figure*}
    \centering
    \includegraphics[trim={1cm 0 0 0}, clip=true,width=0.9\linewidth]{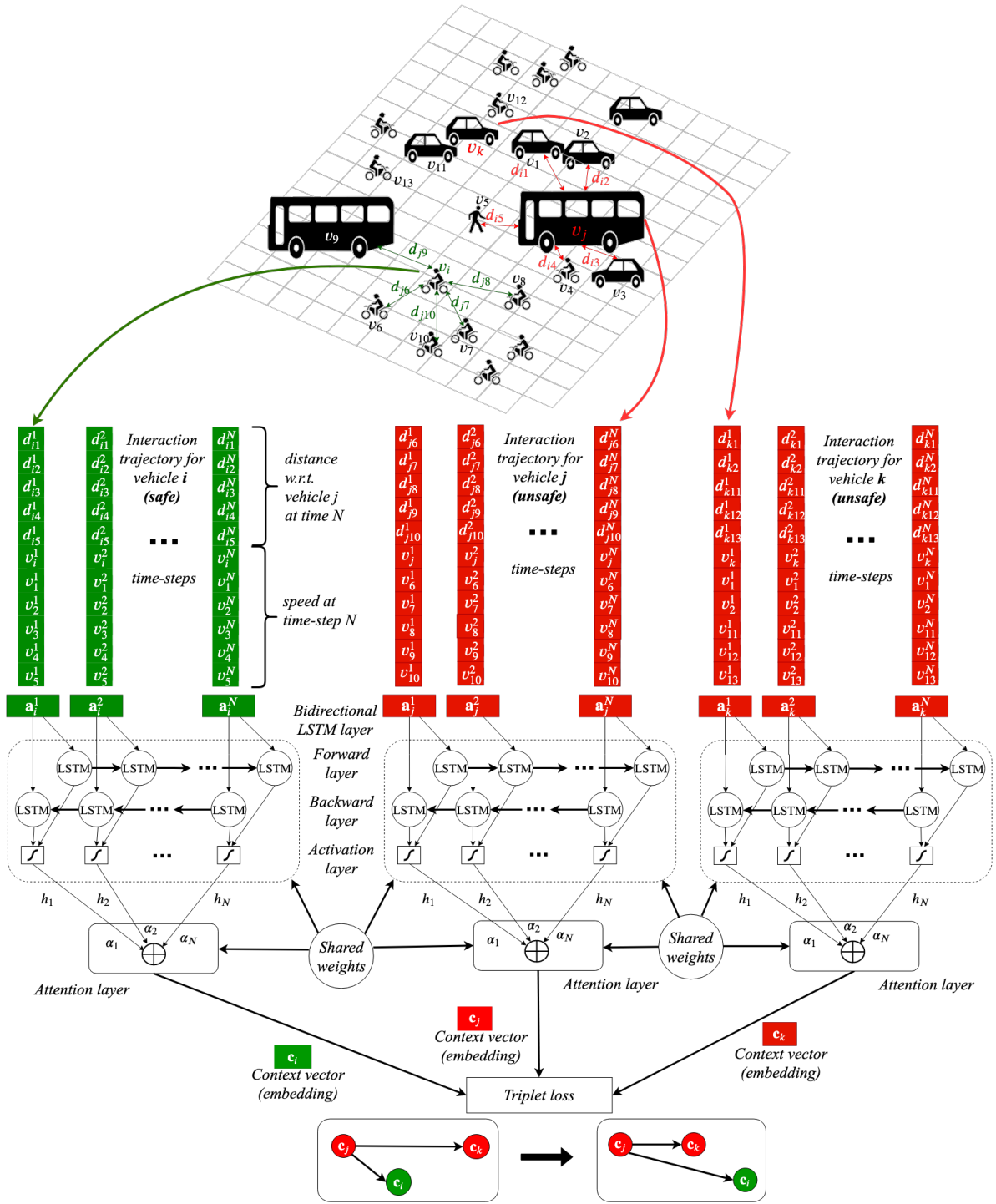}
    \caption{Training of the Siamese Interaction LSTM (SILSTM) using Triplet Loss. The interaction trajectories of three vehicles can be compared based on the neighborhood information at each time-step that is captured using the bidirectional LSTMs and the attention layer. Best viewed in color.}
    \label{SILSTM}
\end{figure*}

In this work, the SILSTM network is built with one or more bidirectional LSTM (BLSTM) layer(s). Each BLSTM layer has Rectified Linear Units (ReLU) activation function at the output of each BLSTM unit. The activation outputs at the ReLU units at each time-step of the final BLSTM layer are pooled to produce a fixed-dimensional output that is sent through an attention layer. Every time-step in the vehicle interaction trajectory is not important in determining the overall safety. There are some crucial interactions in the entire interaction trajectory that are more important than all the others. The attention mechanism assigns a weight $\alpha_n$ to each of the $n$ activation outputs of the BLSTM $h_n$ such that $\sum_{n=1}^N \alpha_n = 1$. The output of the attention layer is the context vector which is calculated by multiplying the attention weight to the hidden output $\mathbf{c} = [\alpha_1 h_1, \alpha_2 h_2, \cdots, \alpha_N h_N]$.
Let $\mathbf{c}_i$, $\mathbf{c}_j$, and $\mathbf{c}_k$ be the context vectors for vehicle interaction trajectories $\mathbf{a}_i$, $\mathbf{a}_j$ and $\mathbf{a}_k$, respectively. Finally, the SILSTM is optimized using triplet loss that is computed as 
\begin{equation}
    \mathcal{L}_{ijk} = max(\| \mathbf{c}_i - \mathbf{c}_j \|^2 - \| \mathbf{c}_i - \mathbf{c}_k \|^2 +m, 0),
\end{equation}
where $\mathbf{c}_i$ and $\mathbf{c}_j$ belong to the same class (either safe or unsafe) and $\mathbf{c}_k$ belongs to a different class. The triplet loss is minimized so that distance between the context vectors from the same class ($\| \mathbf{c}_i - \mathbf{c}_j \|^2$) is pushed to 0 and the distance between context vectors from different classes ($ \| \mathbf{c}_i - \mathbf{c}_k \|^2$) is made to be greater than $\| \mathbf{c}_i - \mathbf{c}_j \|^2 + m$, where $m$ is the margin. A pictorial description of the entire SILSTM network with triplet loss based training is shown in Figure \ref{SILSTM}.

\section{Experimental Evaluation}\label{experiment}
We describe the various experimental details such as the dataset, parameter settings, and protocols in this section. Also, we present and discuss the various quantitative and qualitative results obtained from the experiments. 

\subsection{Dataset}
\begin{figure}
    \centering
    \begin{tabular}{c}
      \includegraphics[width=0.8\linewidth,trim={0 1.2cm 0 1.2cm}, clip = true]{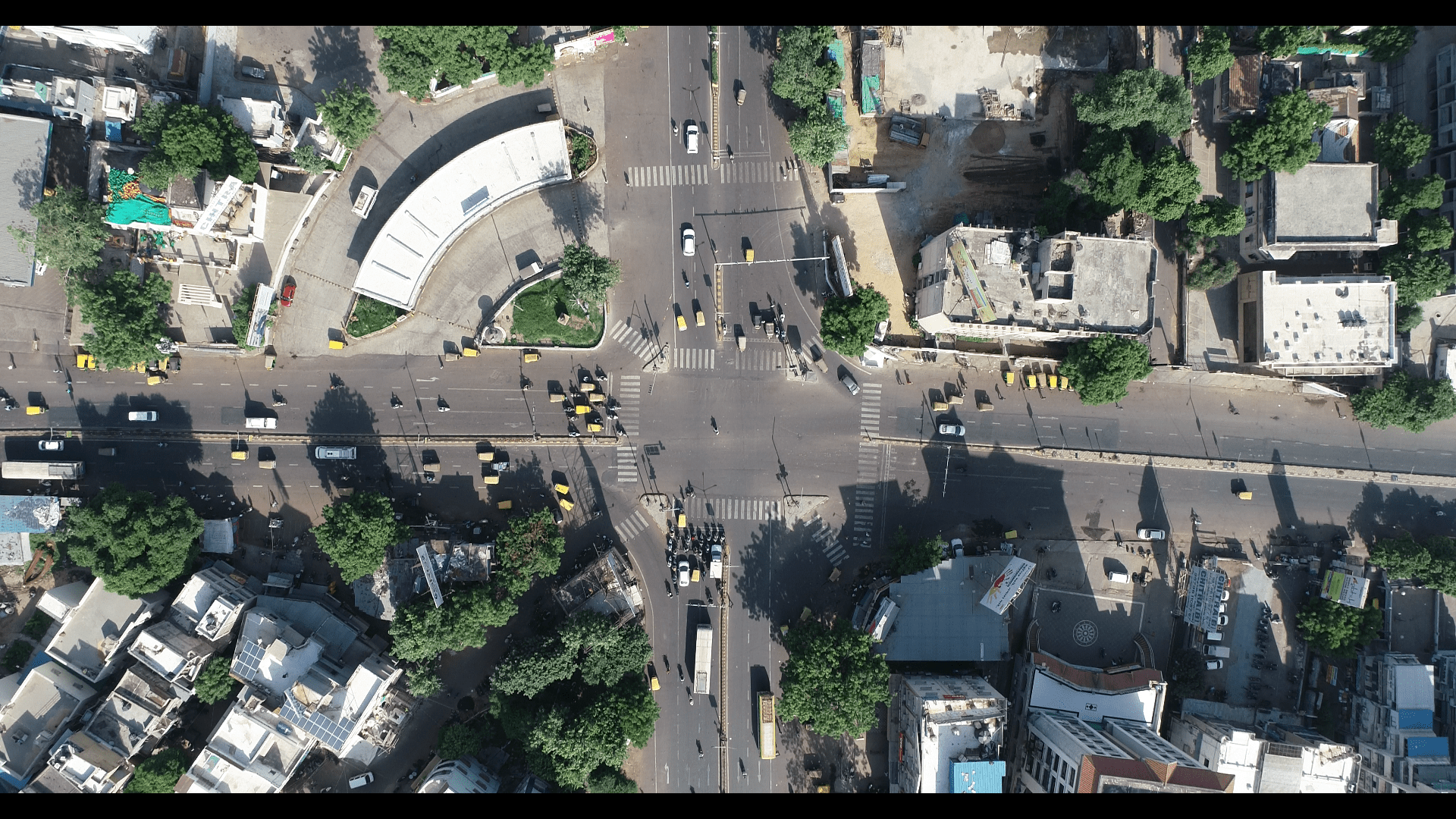} \\
      (a) Paldi (P)\\
      \includegraphics[width=0.8\linewidth,trim={0 1.2cm 0 1.2cm}, clip = true]{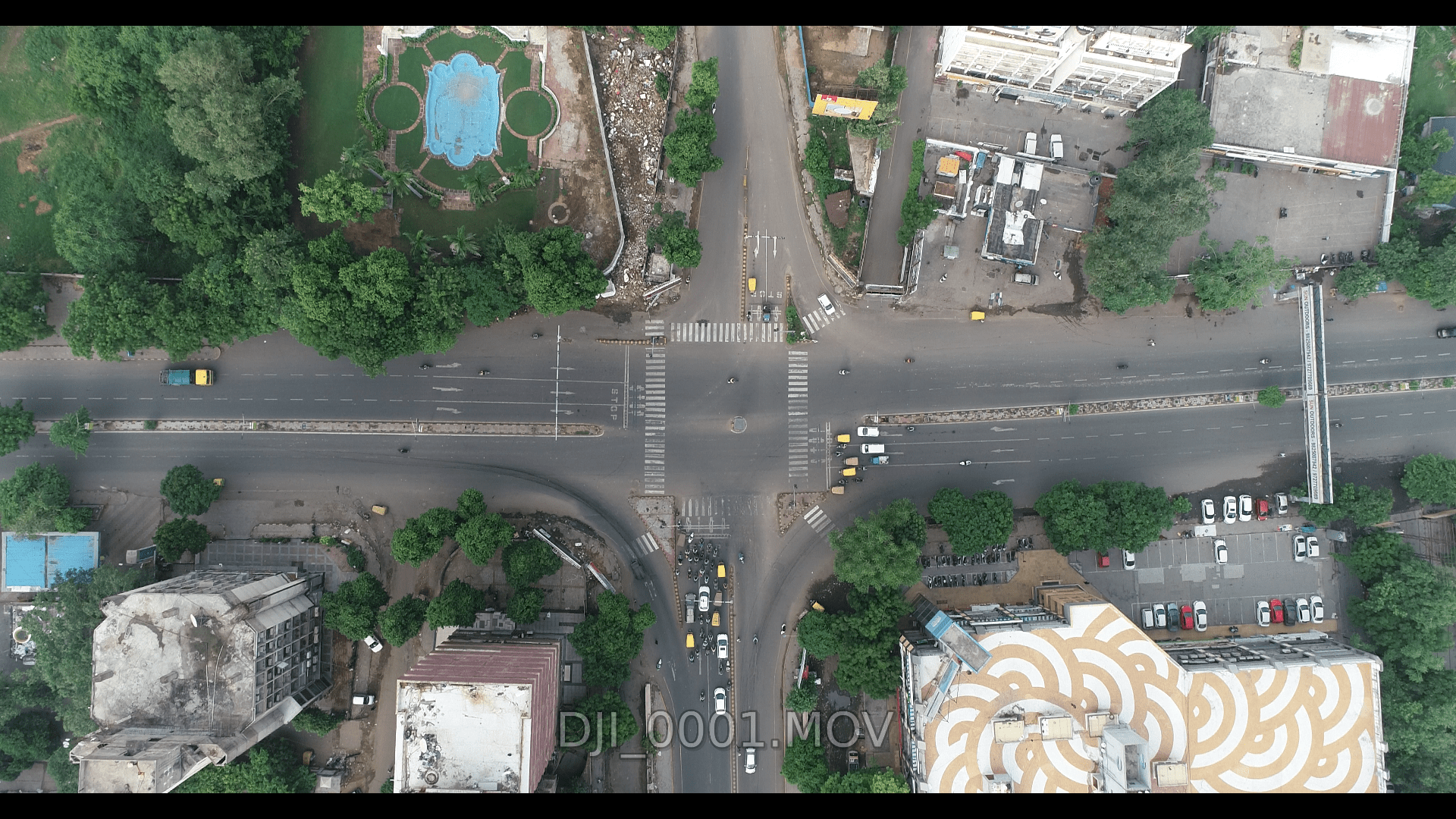} \\
      (b)Nehru bridge - Ashram road (N) \\
      \includegraphics[width=0.8\linewidth,trim={0 1.2cm 0 1.2cm}, clip = true]{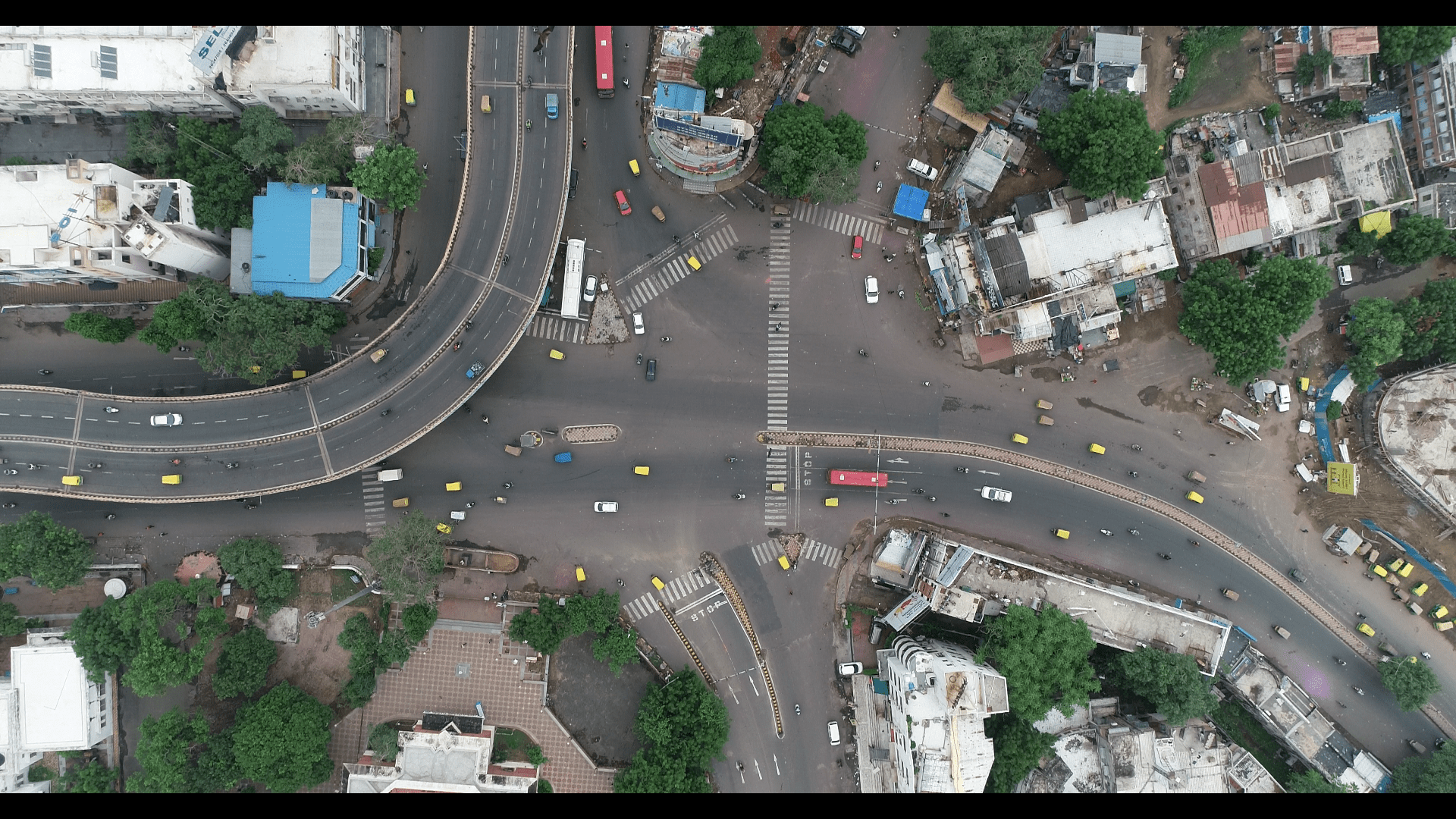} \\
      (c) Swami Vivekananda bridge - Ashram road (V)\\
      \includegraphics[width=0.8\linewidth,trim={0 1.2cm 0 1.2cm}, clip = true]{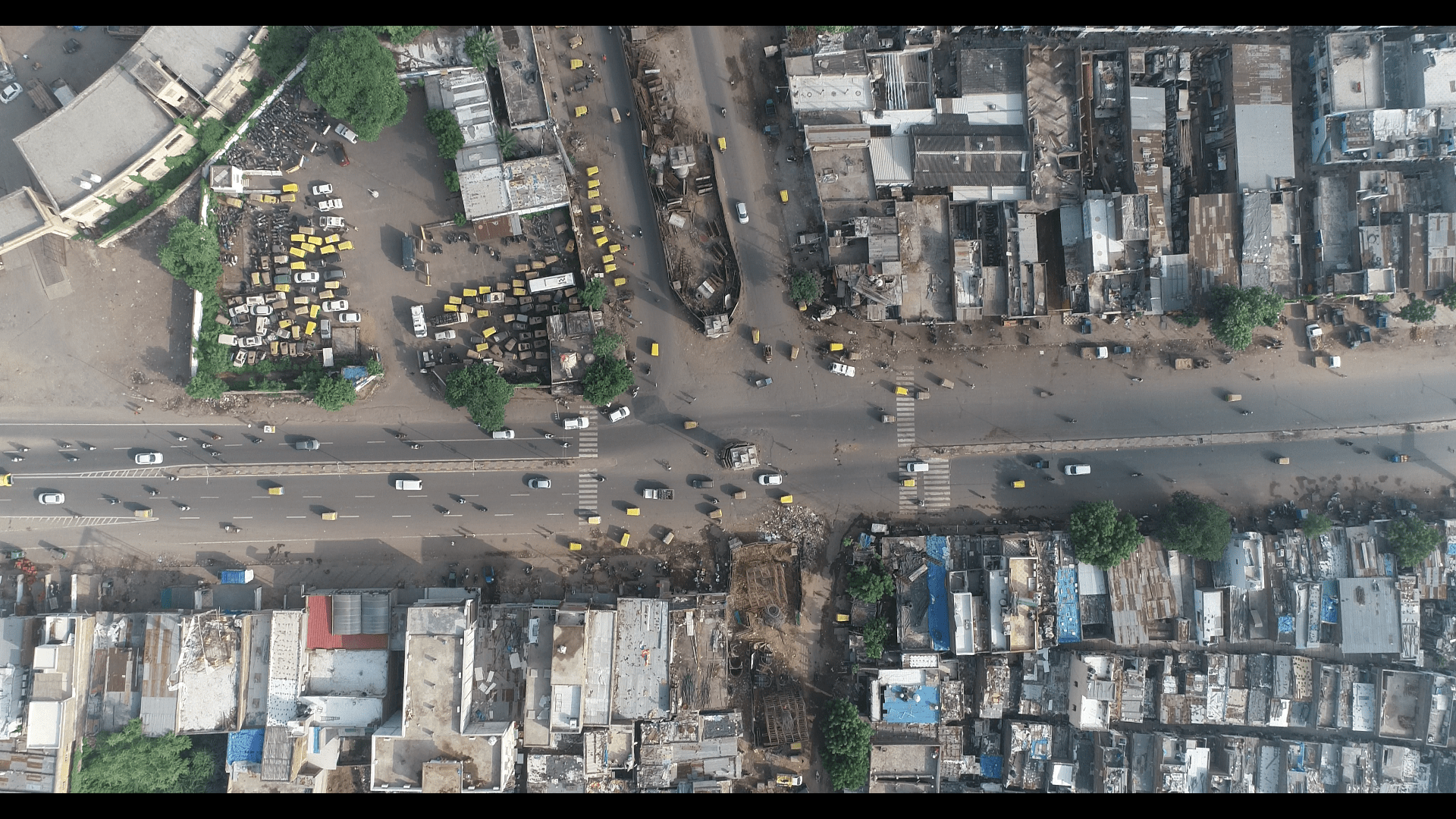} \\
      (d) APMC market (A)\\
    \end{tabular}
    \caption{Intersections in the SkyEye dataset at Ahmedabad in India with their initials in brackets which is used in the rest of the paper. Best viewed in color.}
    \label{intersections}
\end{figure}

\textit{SkyEye:} The SkyEye dataset is the first aerial dataset for monitoring intersections with mixed traffic and lane-less behavior. Around 1 hour of video each from 4 intersections, namely, \textit{Paldi (P), Nehru bridge - Ashram road (N), Swami Vivekananda bridge - Ashram road (V),} and \textit{APMC market (A)} in the city of Ahmedabad, India as shown in Figure \ref{intersections}. These intersections were considered because of the diverse traffic conditions they present. While  \textit{Paldi} and \textit{Nehru bridge } are four-way signalized intersections, the intersection at \textit{Swami Vivekananda bridge} is a seven-way signalized intersection, and \textit{APMC market} is a three-way non-signalized intersection. Hence, this dataset comprehensively covers a wide variety of traffic conditions for both signalized and non-signalized intersections. The videos were captured using the included camera in the DJI Phantom 4 Pro drone at 50 frames per second in 4K resolution (4096$\times$2160). The annotated dataset contains 50,000 frames in total from all the intersections. In these 50,000 frames, a total of 4,021 distinct vehicle tracks are annotated that include 421 cars, 77 buses, 2,185 two-wheelers, and 973 auto-rickshaws. The annotation of these vehicle interaction trajectories as safe and unsafe is discussed in the next subsection.


\subsection{Labelling collision-prone trajectories using Collision Energy}
Annotating collision prone vehicle interaction trajectories is highly subjective. Hence, we propose an annotation scheme based on objective assessment of collision potential in terms of collision energy \cite{social_etiquette}. Collision energy is defined as
\begin{equation}\label{social_sensitivity}
E_{c}(\mathbf{v}; s_i, \mathbf{s}_{j \neq i}| \sigma_d, \sigma_w, \beta) = \sum_{j \neq i} w(s_i, s_j) \exp \left ( −\frac{d^2(\mathbf{v},s_i,s_j)}{2\sigma^2_d} \right ),
\end{equation}
where 
\begin{equation}
w(s_i,s_j) = exp \left ( -\frac {|\Delta \mathbf{p}_{ij}|}{−2 \sigma_w}\right ) . \left ( \frac{1}{2} \left ( 1 - \frac{\Delta \mathbf{p}_{ij}}{|\Delta \mathbf{p}_{ij}|} \frac{\mathbf{v}_i}{| \mathbf{v}_i |}, \right ) \right )^ \beta \end{equation}
and 
\begin{equation}
    d^2 (\mathbf{v}, s_i, s_j ) = \left | \Delta \mathbf{p}_{ij} - \frac{\Delta \mathbf{p}_{ij} (\mathbf{v} - \mathbf{v}_j)}{|\mathbf{v} - \mathbf{v}_j |^2}(\mathbf{v} - \mathbf{v}_j)\right |.
\end{equation}
Here, $\sigma_d$ is the preferred distance a vehicle maintains with each surrounding vehicle to avoid collision, $\sigma_w$ is the distance at which a vehicle reacts to prevent a collision while overtaking, merging, or avoiding oncoming traffic, and $\beta$ is the peakiness of the weighting function for turning distance. In Equation \ref{social_sensitivity}, vehicle $i$ is defined by a state variable $s_i = \{\mathbf{p}_i , \mathbf{v}_i \}$, where $\mathbf{p}_i =(x_i,y_i)$ is the position, and $\mathbf{v}_i$ the velocity of the vehicle. Also, $\Delta \mathbf{p}_{ij}$ denotes the distance between vehicles $i$ and $j$.

As the goal is for all vehicles to navigate in the same space without collisions, we can obtain the parameters $\sigma_d$, $\sigma_w$, and $\beta$ by minimizing collision potential for every vehicle as follows
\begin{equation}
\begin{split}    
\{\sigma_d(i), & \sigma_w(i), \beta(i)\} = \\
    & \argminA_{\{\sigma_d(i), \sigma_w(i), \beta(i)\}} \left ( E_{c}(\mathbf{v}_i; s_i, \mathbf{s}_{-i}| \sigma_d(i), \sigma_w(i), \beta(i))\right ).
\end{split}
\end{equation}
As there are thousands of vehicle interaction trajectories for which the above minimization problem needs to be solved, a fast solver is desirable. Hence, we formulate the above minimization as a genetic algorithm problem instead of the interior point method used in \cite{social_etiquette}.
\begin{figure}
    \centering
    \includegraphics[trim={0.5cm 0.9cm 0.5cm 1cm},clip=true,width=\linewidth]{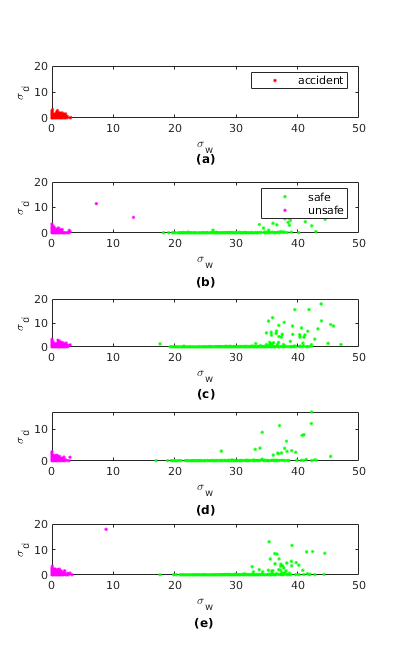}
    \caption{Labelling safe and unsafe trajectories in SkyEye based on turning distance $\sigma_w$ (x-axis) and preferred distance $\sigma_d$ (y-axis) in each figure. (a) The vehicle interaction trajectories involved in accidents from the CADP dataset have low turning and preferred distance. Hence, the cluster with low turning distance at various intersections in SkyEye (b) P, (c) V, (d) N, and (e) A, is labeled as unsafe (each point represents a vehicle trajectory). The cluster whose members have a higher value of turning distance is labeled as safe because the neighboring vehicles are further apart. Best viewed in color.}
    \label{block_labeling}
\end{figure}
After obtaining the $\sigma_d$  and $\sigma_w$ values for all the vehicles, we can label the safe and unsafe vehicle interaction trajectories. Figure \ref{block_labeling} shows that trajectories from accidents in CADP dataset \cite{cadp} have low values of $\sigma_d$ and $\sigma_w$.  This means that accident prone vehicles have low $\sigma_d$ and $\sigma_w$. Considering the CADP dataset as the baseline, the cluster of trajectories in SkyEye with low values of $\sigma_d$ and $\sigma_w$ are labeled as collision-prone. The cluster with high values of $\sigma_d$ and $\sigma_w$ is labeled safe as it comprises of vehicles that maintain a safe distance while driving alongside other vehicles and during overtaking and merging, respectively. Almost all the points can be clearly identified as either safe or unsafe but one or two outliers remain that establish the effectiveness of this labeling scheme over a subjective assessment.

Among the 4,021 unique vehicle interaction trajectories in the SkyEye dataset, 2,041 were labeled as unsafe (collision-prone) and the rest 1,980 were labeled as safe. A breakdown by intersection is presented in Table \ref{labeling_breakdown}. The number of unsafe interaction trajectories are comparable to the safe interaction trajectories for all the intersections.  The labeled interaction trajectories form the ground-truth for our collision prone trajectory detection. For training, testing, and validation, the labeled interaction trajectories were randomly split into 70\%, 20\%, and 10\%, respectively. This process was repeated three times to obtain 3 different splits and the results reported here are averaged over the 3 splits. 
\begin{table}[]
    \centering
       \caption{Intersection-wise distribution of labeled unsafe and safe trajectories in the SkyEye dataset.}
    \label{labeling_breakdown}
    \begin{tabular}{|c|c|c|}
\hline
\textbf{Intersection} & \textbf{unsafe} & \textbf{safe} \\ \hline
P & 927 & 901 \\ \hline
V & 482 & 497\\ \hline
N & 223 & 251 \\ \hline
A & 409 & 331 \\ \hline
\textbf{Total} & 2041 & 1980\\ \hline
\end{tabular}
\end{table}
\subsection{Collision prone Trajectory Prediction}
The median length of the interaction trajectory in the SkyEye dataset was found to be 108 and hence, the number of BLSTM units in the SILSTM network was set to 64. To the BLSTM layer, an attention layer of 32 units was added and this SILSTM network was called BLSTM1L+A, where 1L represents the single BLSTM layer and +A represents the attention layer. For the BLSTM layer, the recurrent and activation dropout values were both set to 0.5, and the attention layer dropout was set to 0.1. These values were obtained empirically by cross-validation. The BLSTM1L+A network was trained for 200 epochs with the criteria of triplet loss on the validation data used to save the best model for evaluation. Though triplet loss provides an embedding that separates dissimilar interaction trajectories, it does not allow us to evaluate retrieval performance on test interaction trajectories. Hence, for reporting the recall, precision, and  F1 score of the test interaction trajectories, we used the $k$ nearest neighbor ($k$NN) algorithm. The $k$NN algorithm allows us to determine whether the test interaction trajectory is more close to unsafe or safe interaction trajectories. Out of the three retrieval metrics, \textit{recall} is the most important in measuring the safety of an intersection as it determines how many unsafe interaction trajectories were recovered correctly.

Table \ref{ablation1layer} presents the recall, precision, and F1 scores of unsafe trajectories for the BLSTM1L+A SILSTM network. In order to determine the collision proneness of interaction trajectories, we also evaluated the effect of the number of neighboring vehicles. This is important as dense traffic is encountered at the intersections in the SkyEye dataset and multiple vehicles surround a given vehicle from all directions. For every vehicle, its neighboring vehicles were chosen based on their distance to the vehicle under consideration. From Table \ref{ablation1layer}, it can be observed that considering more than 8 neighbors does not yield better retrieval performance both in terms of recall and F1 score. This can be attributed to the fact that 8 neighbors are enough to cover the immediate vicinity of a vehicle. Considering more neighbors includes vehicles which do not contribute significantly to the driving behavior. 

\begin{table}[]
    \centering
        \caption{Retrieval metrics for unsafe interaction trajectories using BLSTM1L+A SILSTM on the SkyEye dataset. Considering more than 8 neighbors does not yield better performance.}
    \label{ablation1layer}
   \begin{tabular}{|c|l|l|l|}
\hline
    \multicolumn{1}{|l|}{\textbf{\begin{tabular}[c]{@{}l@{}}Neighboring\\ Vehicles\end{tabular}}} & \textbf{Recall} & \textbf{Precision} & \textbf{F1 score} \\ \hline
3 & 0.76 & 0.51 & 0.61 \\ \hline
4 & 0.76 & 0.49 & 0.57 \\ \hline
5 & 0.77 & \textbf{0.52} & 0.62 \\ \hline
6 & 0.80 & 0.49 & 0.60 \\ \hline
7 & 0.80 & 0.48 & 0.60 \\ \hline
8 & \textbf{0.81} & 0.51 & \textbf{0.63} \\ \hline
9 & 0.79 & 0.49 & 0.61 \\ \hline
10 & 0.75 & 0.50 & 0.60 \\ \hline
\end{tabular}
\end{table}

In literature \cite{neculoiu2016learning}, stacked BLSTM networks have been used for better semantic representation of sequences compared to single-layer BLSTM networks. Hence, a stacked 2-layer SILSTM network called BLSTM2L+A was constructed with 64 and 32 BLSTM units in the first and second layer, respectively and connected to a 32-unit attention layer. In Table \ref{ablation2layer}, the retrieval performance of unsafe trajectories with the BLSTM2L+A SILSTM network is presented. Interestingly, the best retrieval performance was again observed for 8 neighbors which follows the behavior of the BLSTM1L+A network. However, the addition of a BLSTM layer improves the highest recall value to 0.84 over 0.81 for the BLSTM1L+A network. The reason for the improved performance is that both the BLSTM layers operate at different timescales. In effect, aggregation of events over different timescales in interaction trajectories allows for a hierarchical representation that can better detect unsafe driving behavior. In order to achieve better performance, we tried to train a 3-layer network (with 64, 32, and 16 BLSTM units) but very low retrieval performance was observed. This is because there is not enough hierarchical information in intersection trajectories that can be better represented using a 3-layer stacked network compared to a 2-layer stacked network. 

\begin{table}[]
    \centering
        \caption{Retrieval metrics for unsafe interaction trajectories using BLSTM2-A SILSTM on the SkyEye dataset. Considering more than 8 neighbors does not yield better performance.}
    \label{ablation2layer}
   \begin{tabular}{|c|l|l|l|}
\hline
    \multicolumn{1}{|l|}{\textbf{\begin{tabular}[c]{@{}l@{}}Neighboring\\ Vehicles\end{tabular}}} & \textbf{Recall} & \textbf{Precision} & \textbf{F1 score} \\ \hline
3 & 0.36 & 0.54 & 0.45 \\ \hline
4 & 0.75 & 0.51 & 0.61 \\ \hline
5 & 0.75 & 0.52 & 0.61 \\ \hline
6 & 0.79 & 0.49 & 0.60 \\ \hline
7 & 0.79 & 0.49 & 0.60 \\ \hline
8 &\textbf{0.84} & \textbf{0.56} & \textbf{0.66} \\ \hline
9 & 0.81 & 0.50 & 0.61 \\ \hline
10 & 0.75 & 0.50 & 0.60 \\ \hline
\end{tabular}
\end{table}

\subsection{Comparison with different architectures}
The existing methods in literature are not designed to analyze the complete trajectory of an individual vehicle. The focus is only on detecting accidents  \cite{accident_anomaly1,accident_anomaly2,accident_anomaly3,tracking_accident} or evaluating the entire scene consisting of multiple vehicles simultaneously \cite{anticipating1}. In our proposed approach, we evaluate the driving style of each vehicle individually. Hence, in this paper, we compare the performance of different variants of the SILSTM network based on the BLSTM2L(+A) architecture. These variants include - a) 2-layer LSTM network (LSTM2L), b) 2-layer LSTM network with attention (LSTM2L+A), c) 2-layer gated recurrent unit (GRU2L), d) 2 GRU layers with 1 attention layer (GRU2L+A) and e) a 2-layer BLSTM network (BLSTM2L). The number of units in each of these variants is kept the same as the BLSTM2L/BLSTM2L+A SILSTM networks. 

According to Table \ref{comparevariants}, the LSTM2L, LSTM2L+A, GRU2L, GRU2L+A, and BLSTM2L networks also demonstrate the best retrieval performance for 8 neighbors and increasing the number of neighbors affects the performance adversely. The GRU units have two gates - reset and update with no memory units and hence are computationally less expensive. For interaction trajectories, the GRU2L+A network performs similar to the BLSTM2L network. Hence, if a computationally inexpensive network is desired to extract local structure in interaction trajectories, GRU units can be used instead of BLSTM units with only marginal loss in recall performance. Furthermore, the attention layer always shows improvement when used with either GRU, LSTM, or BLSTM units. This shows that aggregation of local structure in interaction trajectories is effective for comparison of interaction trajectories. The local structure arises from the small regions of the interaction trajectory, where the probability of collision is high. Comparing these regions is essential to the detection of similarity in collision-prone interaction trajectories.  

\begin{table}[]
    \centering
        \caption{Comparison of retrieval performance for different architectures for the SILSTM network.}
    \label{comparevariants}
\begin{tabular}{|c|c|c|c|c|}
\hline
\multicolumn{1}{|c|}{\textbf{Architecture}} & \textbf{\# Neighbors} & \textbf{Recall} & \textbf{Precision} & \textbf{F1 score} \\ \hline
\multirow{3}{*}{LSTM2L} & 7 & 0.79 & 0.49 & 0.61 \\ \cline{2-5} 
 & 8 & 0.82 & 0.51 & 0.63 \\ \cline{2-5} 
 & 9 & 0.80 & 0.50 & 0.61 \\ \hline
\multirow{3}{*}{LSTM2L+A} & 7 & 0.79 & 0.49 & 0.60 \\ \cline{2-5} 
 & 8 & 0.84 & 0.52 & 0.64 \\ \cline{2-5} 
 & 9 & 0.82 & 0.50 & 0.62 \\ \hline
\multirow{3}{*}{GRU2L} & 7 & 0.76 & 0.49 & 0.57 \\ \cline{2-5} 
 & 8 & 0.80 & 0.51 & 0.61 \\ \cline{2-5} 
 & 9 & 0.78 & 0.48 & 0.59 \\ \hline
\multirow{3}{*}{GRU2L+A} & 7 & 0.77 & 0.51 & 0.59 \\ \cline{2-5} 
& 8 & 0.81 & 0.53 & 0.62 \\ \cline{2-5} 
& 9 & 0.79 & 0.49 & 0.61 \\ \hline
\multirow{3}{*}{BLSTM2L} & 7 & 0.78 & 0.48 & 0.59 \\ \cline{2-5}
& 8 & 0.81 & 0.49 & 0.63 \\ \cline{2-5}
& 9 & 0.80 & 0.48 & 0.61 \\ \hline \hline
{\begin{tabular}[c]{@{}c@{}}\textbf{BLSTM2L+A}\end{tabular}} & \textbf{8} & \textbf{0.84} & \textbf{0.56} & \textbf{0.66} \\ \hline
\end{tabular}
\end{table}

\subsection{Qualitative Analysis}
\begin{figure*}[htbp]
    \centering
    \begin{tabular}{cc}
         \includegraphics[height=200pt,width=0.48\linewidth]{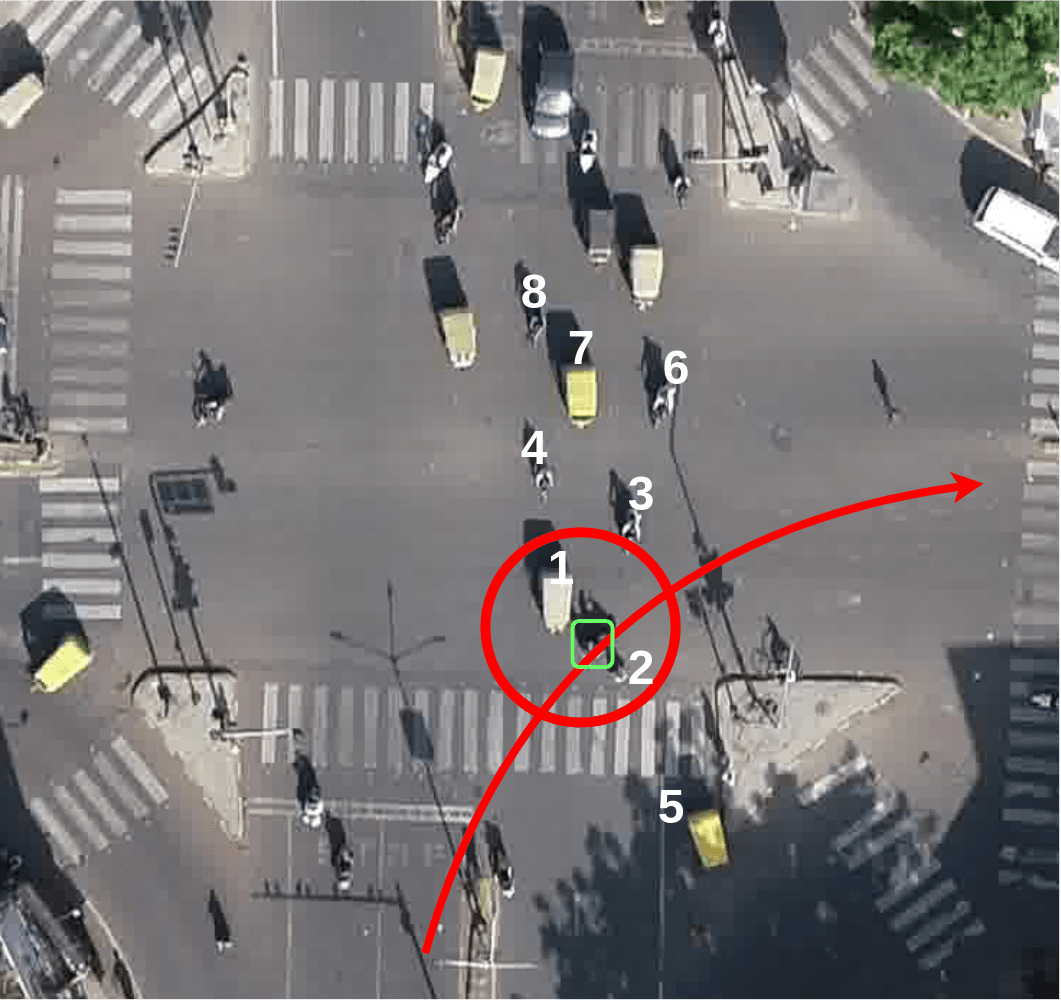}&
         \includegraphics[height=200pt,width=0.48\linewidth]{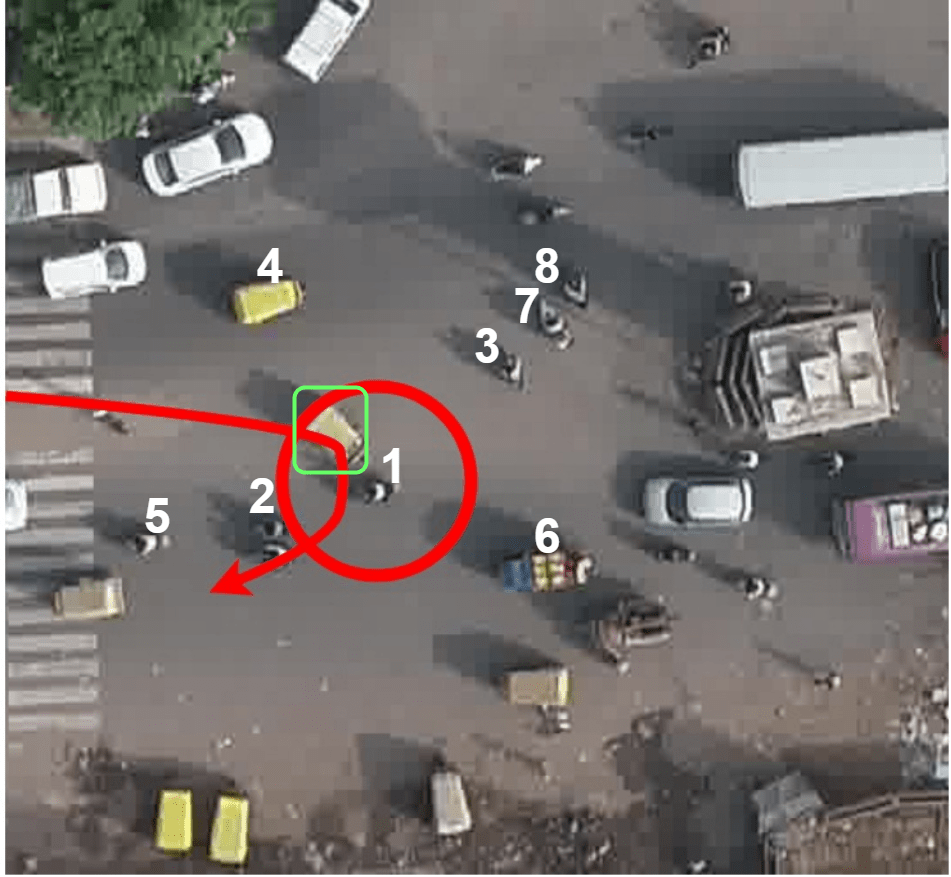}\\
         (a) & (b) \\
         \includegraphics[height=200pt,width=0.48\linewidth]{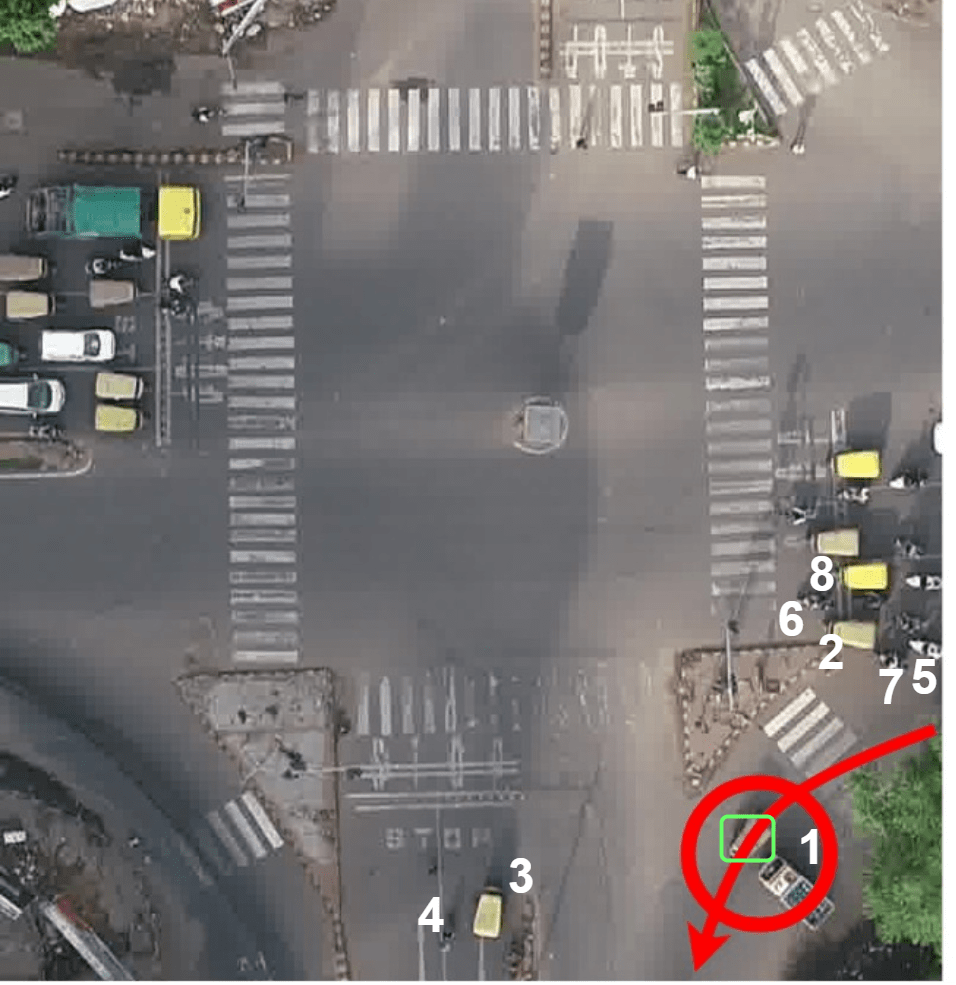}&
         \includegraphics[height=200pt,width=0.48\linewidth]{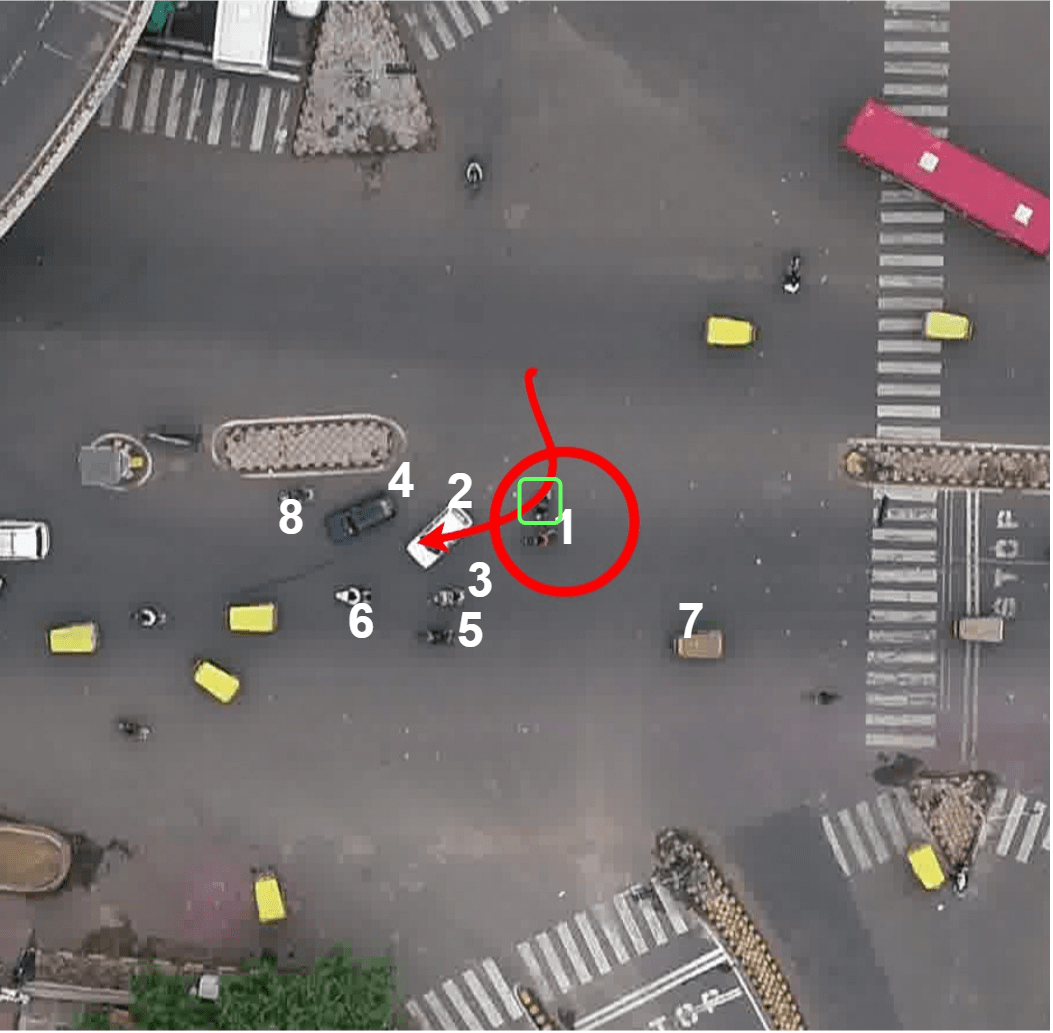}\\
         (c) & (d)
    \end{tabular}
    \caption{Examples of unsafe interaction trajectories in SkyEye dataset at various intersections - (a) P, (b) A, (c) N, and (d) V. The vehicle whose interaction trajectory is under consideration is shown in green and its neighbors for that particular instant are shown with numbers in white. The trajectory is shown in red and an unsafe interaction is marked in red.}
    \label{visualsilstm}
\end{figure*}

Some examples of the detected unsafe/collision-prone interaction trajectories using the SILSTM network at the 4 intersections of the SkyEye dataset are shown in Figure \ref{visualsilstm}. The vehicle whose trajectory is under consideration is shown (in green) with its 8 nearest neighbors for that particular instant also marked with numbers (in white). In each of these cases, one particular instance is highlighted along the trajectory of the vehicle where a probable collision is about to happen with one of the neighbors. Such an collision-prone interaction contributes to the unsafe nature of the vehicle interaction trajectory. Most of these unsafe vehicle interactions occur when a vehicle emerges in a direction opposite to the prevalent flow of the traffic. As the prevalent flow has considerable speed compared to the emerging vehicle, there is a imminent chance of collision that is identified by SILSTM when analyzing the interaction trajectory.

\subsection{Quantitative Analysis}
For the different variants of the SILSTM network, a high recall rate is observed with relatively lower precision values. A low precision value indicates many false positives that arise because many benign intersection trajectories are considered as collision-prone. As the SILSTM method considers vehicle speed in addition to distance for modeling intersections, the intersections with smaller vehicles rapidly traversing between large stationary vehicles at stop signs are also considered as misclassified as unsafe. In Table \ref{intersectionwise}, the recall, precision, and F1 values for each intersection in the SkyEye dataset are presented separately. The two intersections, namely, Paldi (P) and APMC market (A) show much higher recall values compared to Swami Vivekananda bridge - Ashram road (V) and Nehru Bridge - Ashram Road (N). This means that unsafe interaction trajectories are misclassified less at P and A intersections but are misclassified the most at intersection V. As intersection V is a 7-way signalized intersection, there are a number of concurrent traffic flows. Based on the distance and high relative speed between the various concurrent traffic flows, many safe interaction trajectories are misclassified as unsafe. 

\begin{table}[htbp]
\centering
\caption{Retrieval metrics for every intersection in SkyEye dataset based on the best performing BLSTM2L+A SILSTM network considering 8 neighbors for each vehicle. }
\label{intersectionwise}
\begin{tabular}{|c|c|c|c|}
\hline
\textbf{Intersection} & \multicolumn{1}{l|}{\textbf{Recall}} & \multicolumn{1}{l|}{\textbf{Precision}} & \multicolumn{1}{l|}{\textbf{F1 score}} \\ \hline
P & 0.90 & 0.52 & 0.66 \\ \hline
V & 0.51 & 0.43 & 0.47 \\ \hline
N & 0.61 & 0.50 & 0.55 \\ \hline
A & 1.00 & 0.54 & 0.70 \\ \hline
\end{tabular}
\end{table}

\section{Conclusion} \label{conclusion}
In this paper, we have proposed a Siamese Interaction Long Short-Term Memory network (SILSTM) that can compare the driving style of a vehicle with another vehicle based on interactions with neighboring vehicles. The interactions were represented in the form of interaction trajectories that contained the distance of a vehicle from its neighbors and the speed of the neighbors. The proposed SILSTM quantitatively identifies unsafe vehicle interaction trajectories at different types of intersections in challenging lane-less traffic conditions. Also, a large aerial dataset called SkyEye was introduced that is the first to provide long-term monitoring of signalized/non-signalized intersections with lane-less traffic in India. We demonstrated the efficacy of the proposed SILSTM approach in learning salient features from long vehicle interaction trajectories. We showed that learning these salient  features allowed for highly effective detection of collision-prone trajectories at various types of intersections in the SkyEye dataset.

\section*{Acknowledgement}
This work has been conducted as the part of SATREPS project entitled on “Smart Cities development for Emerging Countries by Multimodal Transport System based on Sensing, Network and Big Data Analysis of Regional Transportation” (JPMJSA1606) funded by JST and JICA.


\begin{thebibliography}{10}
\providecommand{\url}[1]{#1}
\csname url@samestyle\endcsname
\providecommand{\newblock}{\relax}
\providecommand{\bibinfo}[2]{#2}
\providecommand{\BIBentrySTDinterwordspacing}{\spaceskip=0pt\relax}
\providecommand{\BIBentryALTinterwordstretchfactor}{4}
\providecommand{\BIBentryALTinterwordspacing}{\spaceskip=\fontdimen2\font plus
\BIBentryALTinterwordstretchfactor\fontdimen3\font minus
  \fontdimen4\font\relax}
\providecommand{\BIBforeignlanguage}[2]{{%
\expandafter\ifx\csname l@#1\endcsname\relax
\typeout{** WARNING: IEEEtran.bst: No hyphenation pattern has been}%
\typeout{** loaded for the language `#1'. Using the pattern for}%
\typeout{** the default language instead.}%
\else
\language=\csname l@#1\endcsname
\fi
#2}}
\providecommand{\BIBdecl}{\relax}
\BIBdecl

\bibitem{intersection_crash_report}
N.~H. T.~S. Administration \emph{et~al.}, ``Crash factors in
  intersection-related crashes: An on-scene perspective,'' \emph{Nat. Center
  Stat. Anal., National Highway Traffic Safety Administration, Washington, DC,
  USA, Tech. Rep. DOT HS}, vol. 811366, 2010.

\bibitem{roadsafetyreview}
M.~P. Hagenzieker, J.~J. Commandeur, and F.~D. Bijleveld, ``The history of road
  safety research: A quantitative approach,'' \emph{Transportation research
  part F: traffic psychology and behaviour}, vol.~25, pp. 150--162, 2014.

\bibitem{statistical1}
F.~Prieto, E.~Gómez-Déniz, and J.~M. Sarabia, ``Modelling road accident
  blackspots data with the discrete generalized pareto distribution,''
  \emph{Accident Analysis and Prevention}, vol.~71, pp. 38 -- 49, 2014.

\bibitem{statistical2}
A.~Azadeh, M.~Zarrin, and M.~Hamid, ``A novel framework for improvement of road
  accidents considering decision-making styles of drivers in a large
  metropolitan area,'' \emph{Accident Analysis and Prevention}, vol.~87, pp. 17
  -- 33, 2016.

\bibitem{blackspot1}
M.~Ghadi and A.~Torok, ``A comparative analysis of black spot identification
  methods and road accident segmentation methods,'' \emph{Accident Analysis and
  Prevention}, vol. 128, pp. 1 -- 7, 2019.

\bibitem{blackspot2}
B.~Debrabant, U.~Halekoh, W.~H. Bonat, D.~L. Hansen, J.~Hjelmborg, and
  J.~Lauritsen, ``Identifying traffic accident black spots with poisson-tweedie
  models,'' \emph{Accident Analysis and Prevention}, vol. 111, pp. 147 -- 154,
  2018.

\bibitem{blackspot3}
A.~Sandhyavitri, Zamri, S.~Wiyono, and Subiantoro, ``Three strategies reducing
  accident rates at black spots and black sites road in riau province,
  indonesia,'' \emph{Transportation Research Procedia}, vol.~25, pp. 2153 --
  2166, 2017, world Conference on Transport Research - WCTR 2016 Shanghai.
  10-15 July 2016.

\bibitem{blackspot4}
M.~Ghadi and A.~Torok, ``Comparison of different black spot identification
  methods,'' \emph{Transportation Research Procedia}, vol.~27, pp. 1105 --
  1112, 2017, 20th EURO Working Group on Transportation Meeting, EWGT 2017, 4-6
  September 2017, Budapest, Hungary.

\bibitem{drivingsurvey}
C.~{Marina Martinez}, M.~{Heucke}, F.~{Wang}, B.~{Gao}, and D.~{Cao}, ``Driving
  style recognition for intelligent vehicle control and advanced driver
  assistance: A survey,'' \emph{IEEE Transactions on Intelligent Transportation
  Systems}, vol.~19, no.~3, pp. 666--676, March 2018.

\bibitem{ua-detrac}
S.~Lyu, M.-C. Chang, D.~Du, L.~Wen, H.~Qi, Y.~Li, Y.~Wei, L.~Ke, T.~Hu,
  M.~Del~Coco \emph{et~al.}, ``Ua-detrac 2017: Report of avss2017 \& iwt4s
  challenge on advanced traffic monitoring,'' in \emph{Advanced Video and
  Signal Based Surveillance (AVSS), 2017 14th IEEE International Conference
  on}.\hskip 1em plus 0.5em minus 0.4em\relax IEEE, 2017, pp. 1--7.

\bibitem{datafromsky}
J.~Apeltauer, A.~Babinec, D.~Herman, and T.~Apeltauer, ``Automatic vehicle
  trajectory extraction for traffic analysis from aerial video data,''
  \emph{The International Archives of Photogrammetry, Remote Sensing and
  Spatial Information Sciences}, vol.~40, no.~3, p.~9, 2015.

\bibitem{social_mixed}
D.~N. Huynh, M.~Boltze, and A.~T. Vu, ``Modelling mixed traffic flow at
  signalized intersection using social force model,'' \emph{Journal of the
  Eastern Asia Society for Transportation Studies}, vol.~10, pp. 1734--1749,
  2013.

\bibitem{social2d}
W.~Huang, M.~Fellendorf, and R.~Sch{\"o}nauer, ``Social force based vehicle
  model for 2-dimensional spaces,'' in \emph{91st Annual Meeting of the
  Transportation Research Board. Washington, DC, USA}, 2011.

\bibitem{convolutional_social}
N.~Deo and M.~M. Trivedi, ``Convolutional social pooling for vehicle trajectory
  prediction,'' in \emph{The IEEE Conference on Computer Vision and Pattern
  Recognition (CVPR) Workshops}, June 2018.

\bibitem{social_lstm}
A.~Alahi, K.~Goel, V.~Ramanathan, A.~Robicquet, L.~Fei-Fei, and S.~Savarese,
  ``Social lstm: Human trajectory prediction in crowded spaces,'' in
  \emph{Proceedings of the IEEE Conference on Computer Vision and Pattern
  Recognition}, 2016, pp. 961--971.

\bibitem{social_gan}
A.~Gupta, J.~Johnson, L.~Fei-Fei, S.~Savarese, and A.~Alahi, ``Social gan:
  Socially acceptable trajectories with generative adversarial networks,'' in
  \emph{2018 IEEE/CVF Conference on Computer Vision and Pattern Recognition},
  June 2018, pp. 2255--2264.

\bibitem{social_attention}
A.~Vemula, K.~Muelling, and J.~Oh, ``Social attention: Modeling attention in
  human crowds,'' in \emph{2018 IEEE International Conference on Robotics and
  Automation (ICRA)}.\hskip 1em plus 0.5em minus 0.4em\relax IEEE, 2018, pp.
  1--7.

\bibitem{lanelessbehavior}
G.~Asaithambi, V.~Kanagaraj, and T.~Toledo, ``Driving behaviors: Models and
  challenges for non-lane based mixed traffic,'' \emph{Transportation in
  Developing Economies}, vol.~2, no.~2, p.~19, 2016.

\bibitem{whoareyouwith}
K.~{Yamaguchi}, A.~C. {Berg}, L.~E. {Ortiz}, and T.~L. {Berg}, ``Who are you
  with and where are you going?'' in \emph{CVPR 2011}, June 2011, pp.
  1345--1352.

\bibitem{visdrone}
P.~Zhu, L.~Wen, X.~Bian, L.~Haibin, and Q.~Hu, ``Vision meets drones: A
  challenge,'' \emph{arXiv preprint arXiv:1804.07437}, 2018.

\bibitem{accident_anomaly3}
K.~{Yun}, H.~{Jeong}, K.~M. {Yi}, S.~W. {Kim}, and J.~Y. {Choi}, ``Motion
  interaction field for accident detection in traffic surveillance video,'' in
  \emph{2014 22nd International Conference on Pattern Recognition}, Aug 2014,
  pp. 3062--3067.

\bibitem{accident_anomaly2}
D.~{Singh} and C.~K. {Mohan}, ``Deep spatio-temporal representation for
  detection of road accidents using stacked autoencoder,'' \emph{IEEE
  Transactions on Intelligent Transportation Systems}, vol.~20, no.~3, pp.
  879--887, March 2019.

\bibitem{accident_anomaly1}
W.~Sultani, C.~Chen, and M.~Shah, ``Real-world anomaly detection in
  surveillance videos,'' in \emph{Proceedings of the IEEE Conference on
  Computer Vision and Pattern Recognition}, 2018, pp. 6479--6488.

\bibitem{anticipating1}
F.-H. Chan, Y.-T. Chen, Y.~Xiang, and M.~Sun, ``Anticipating accidents in
  dashcam videos,'' in \emph{Asian Conference on Computer Vision}.\hskip 1em
  plus 0.5em minus 0.4em\relax Springer, 2016, pp. 136--153.

\bibitem{cadp}
A.~P. Shah, J.-B. Lamare, T.~Nguyen-Anh, and A.~Hauptmann, ``Cadp: A novel
  dataset for cctv traffic camera based accident analysis,'' in \emph{2018 15th
  IEEE International Conference on Advanced Video and Signal Based Surveillance
  (AVSS)}.\hskip 1em plus 0.5em minus 0.4em\relax IEEE, 2018, pp. 1--9.

\bibitem{accident_anticipating2}
T.~Suzuki, H.~Kataoka, Y.~Aoki, and Y.~Satoh, ``Anticipating traffic accidents
  with adaptive loss and large-scale incident db,'' in \emph{Proceedings of the
  IEEE Conference on Computer Vision and Pattern Recognition}, 2018, pp.
  3521--3529.

\bibitem{accidenttypes}
N.~Distefano and S.~Leonardi, ``A list of accident scenarios for three legs
  skewed intersections,'' \emph{IATSS research}, vol.~42, no.~3, pp. 97--104,
  2018.

\bibitem{vgg}
K.~Simonyan and A.~Zisserman, ``Very deep convolutional networks for
  large-scale image recognition,'' \emph{CoRR}, vol. abs/1409.1556, 2014.

\bibitem{car_following}
G.~F. Newell, ``A simplified car-following theory: a lower order model,''
  \emph{Transportation Research Part B: Methodological}, vol.~36, no.~3, pp.
  195--205, 2002.

\bibitem{empirical_tri_class}
C.-L. Lan and G.-L. Chang, ``Empirical observations and formulations of
  tri-class traffic flow properties for design of traffic signals,'' \emph{IEEE
  Transactions on Intelligent Transportation Systems}, vol.~20, no.~3, pp.
  830--842, 2018.

\bibitem{original_social_force}
D.~Helbing and P.~Molnar, ``Social force model for pedestrian dynamics,''
  \emph{Physical review E}, vol.~51, no.~5, p. 4282, 1995.

\bibitem{rnn}
A.~{Graves}, M.~{Liwicki}, S.~{Fernández}, R.~{Bertolami}, H.~{Bunke}, and
  J.~{Schmidhuber}, ``A novel connectionist system for unconstrained
  handwriting recognition,'' \emph{IEEE Transactions on Pattern Analysis and
  Machine Intelligence}, vol.~31, no.~5, pp. 855--868, May 2009.

\bibitem{lstm}
\BIBentryALTinterwordspacing
S.~Hochreiter and J.~Schmidhuber, ``Long short-term memory,'' \emph{Neural
  Computation}, vol.~9, no.~8, pp. 1735--1780, 1997. [Online]. Available:
  \url{https://doi.org/10.1162/neco.1997.9.8.1735}
\BIBentrySTDinterwordspacing

\bibitem{bilstm}
A.~Graves and J.~Schmidhuber, ``Framewise phoneme classification with
  bidirectional lstm and other neural network architectures,'' \emph{Neural
  networks}, vol.~18, no. 5-6, pp. 602--610, 2005.

\bibitem{social_etiquette}
A.~Robicquet, A.~Sadeghian, A.~Alahi, and S.~Savarese, ``Learning social
  etiquette: Human trajectory understanding in crowded scenes,'' in
  \emph{European conference on computer vision}.\hskip 1em plus 0.5em minus
  0.4em\relax Springer, 2016, pp. 549--565.

\bibitem{siameseog}
L.~Fei-Fei, R.~Fergus, and P.~Perona, ``One-shot learning of object
  categories,'' \emph{IEEE Transactions on Pattern Analysis and Machine
  Intelligence}, vol.~28, no.~4, pp. 594--611, April 2006.

\bibitem{siameselstm}
J.~Mueller and A.~Thyagarajan, ``Siamese recurrent architectures for learning
  sentence similarity,'' in \emph{Proceedings of the Thirtieth AAAI Conference
  on Artificial Intelligence}.\hskip 1em plus 0.5em minus 0.4em\relax AAAI
  Press, 2016, pp. 2786--2792.

\bibitem{neculoiu2016learning}
P.~Neculoiu, M.~Versteegh, and M.~Rotaru, ``Learning text similarity with
  siamese recurrent networks,'' in \emph{Proceedings of the 1st Workshop on
  Representation Learning for NLP}, 2016, pp. 148--157.

\bibitem{tracking_accident}
S.~{Kamijo}, Y.~{Matsushita}, K.~{Ikeuchi}, and M.~{Sakauchi}, ``Traffic
  monitoring and accident detection at intersections,'' \emph{IEEE Transactions
  on Intelligent Transportation Systems}, vol.~1, no.~2, pp. 108--118, June
  2000.

\end{thebibliography}
\end{document}